\documentclass[10pt, conference]{IEEEtran}
\usepackage{amsmath,amssymb,amsfonts}
\usepackage{graphicx}
\usepackage{pifont}

\usepackage{amsthm}
\usepackage{xcolor}
\usepackage[figuresright]{rotating}

\usepackage{graphicx}
\graphicspath{{/}{fig/}}

\usepackage{array}
\usepackage{textcomp}
\usepackage{xcolor}
\usepackage{multirow}
\usepackage{booktabs}

\usepackage{mathtools}
\usepackage{breqn}
\usepackage{float}

\usepackage{pgfplots}
\pgfplotsset{compat=1.7}
\usepgfplotslibrary{groupplots}

\usepackage{caption}
\usepackage{subcaption}

\usepackage{multirow,tabularx}
\usepackage{hyperref}
\usepackage{flushend}
\usepackage{algorithmic}
\usepackage[vlined, ruled, shortend]{algorithm2e}
\newlength\figureheight
\newlength\figurewidth
\SetAlCapNameFnt{\footnotesize}
\SetAlCapFnt{\footnotesize}

\title{
    Self-Calibrating Anomaly and Change Detection for Autonomous Inspection Robots
}

\author{
    \IEEEauthorblockN{
        \vspace{1em}
        Sahar Salimpour\IEEEauthorrefmark{2},
        Jorge Pe\~na Queralta\IEEEauthorrefmark{2},
        Tomi Westerlund\IEEEauthorrefmark{2}
    }
    \IEEEauthorblockA{
        \normalsize
        \IEEEauthorrefmark{2}\href{https://tiers.utu.fi}{Turku Intelligent Embedded and Robotic Systems (TIERS) Lab, University of Turku, Finland}.\\
        Emails: \textsuperscript{1}\{sahars, jopequ, tovewe\}@utu.fi\\[+6pt]
    }
}

\begin{document}

\maketitle
\thispagestyle{empty}
\pagestyle{empty}

%%%%%%%%%%%%%%%%%%%%%%%%%%%%%%%%%%%%%%%%%%%%%%
%%                                          %%
%%           ABSTRACT AND TITLE             %%
%%                                          %%
%%%%%%%%%%%%%%%%%%%%%%%%%%%%%%%%%%%%%%%%%%%%%%

%%%%%%%%%%%%%%%%%%%%%%%%%%%%%%%%%%%%%%%%%%%%%%
%%                                          %%
%%                ABSTRACT                  %%
%%                                          %%
%%%%%%%%%%%%%%%%%%%%%%%%%%%%%%%%%%%%%%%%%%%%%%

\begin{abstract}%
    \label{sec:abstract}%
    %
%In recent years, a
Automatic detection of visual anomalies and changes in the environment has been a topic of recurrent attention in the fields of machine learning and computer vision over the past decades. A visual anomaly or change detection algorithm identifies regions of an image that differ from a reference image or dataset. 
The majority of existing approaches focus on anomaly or fault detection in a specific class of images or environments, while general-purpose visual anomaly detection algorithms are more scarce in the literature. In this paper, we propose a  comprehensive deep learning framework for detecting anomalies and changes in a priori unknown environments after a reference dataset is gathered, and without need for retraining the model. We use the SuperPoint and SuperGlue feature extraction and matching methods to detect anomalies based on reference images taken from a similar location and with partial overlapping of the field of view. 
We also introduce a self-calibrating method for the proposed model in order to address the problem of sensitivity to feature matching thresholds and environmental conditions. %We then use Mask-RCNN instance segmentation and DBSCAN to cluster unmatched interest points as anomalous regions in images. 
To evaluate the proposed framework, we have used a ground robot system for the purpose of reference and query data collection. We show that high
accuracy can be obtained using the proposed method. We also show that the calibration process enhances changes and foreign object detection performance. 
%The work presented in this paper opens the door to more general deep learning solutions for autonomous inspection robots across different operational environments.

\end{abstract}

\IEEEpeerreviewmaketitle

%%%%%%%%%%%%%%%%%%%%%%%%%%%%%%%%%%%%%%%%%%%%%%
%%                                          %%
%%                SECTIONS                  %%
%%                                          %%
%%%%%%%%%%%%%%%%%%%%%%%%%%%%%%%%%%%%%%%%%%%%%%
%%%%%%%%%%%%%%%%%%%%%%%%%%%%%%%%%%%%%%%%%%%%%%
%%                                          %%
%%              INTRODUCTION                %%
%%                                          %%
%%%%%%%%%%%%%%%%%%%%%%%%%%%%%%%%%%%%%%%%%%%%%%

\section{Introduction}\label{sec:introduction}

Anomaly detection, also known as foreign and outlier detection is a recurrent concept in computer vision, machine learning, and statistics. It has been explored in a wide range of research and application fields such as industry~\cite{kamat2020anomaly}, medical imaging~\cite{fernando2020deep}, security and safety systems~\cite{akcay2018ganomaly}. Anomaly detectors are designed to identify the presence of unknown artifacts in data types such as images, videos, audio, text, and time series that significantly differs from the normal data. Visual anomaly detection refers to the detection of anomalies within image data. Vision-based anomaly detection can be performed on both pixel and image levels. Anomaly detection algorithms are typically based on a reference dataset from which \textit{normal} conditions are generalized. Change detection algorithms, on the other hand, typically compare individual pairs of images to detect changes. Our objective is to combine both from the perspective of supporting autonomous inspection robots.

\begin{figure}
    \centering
    \includegraphics[width=.48\textwidth]{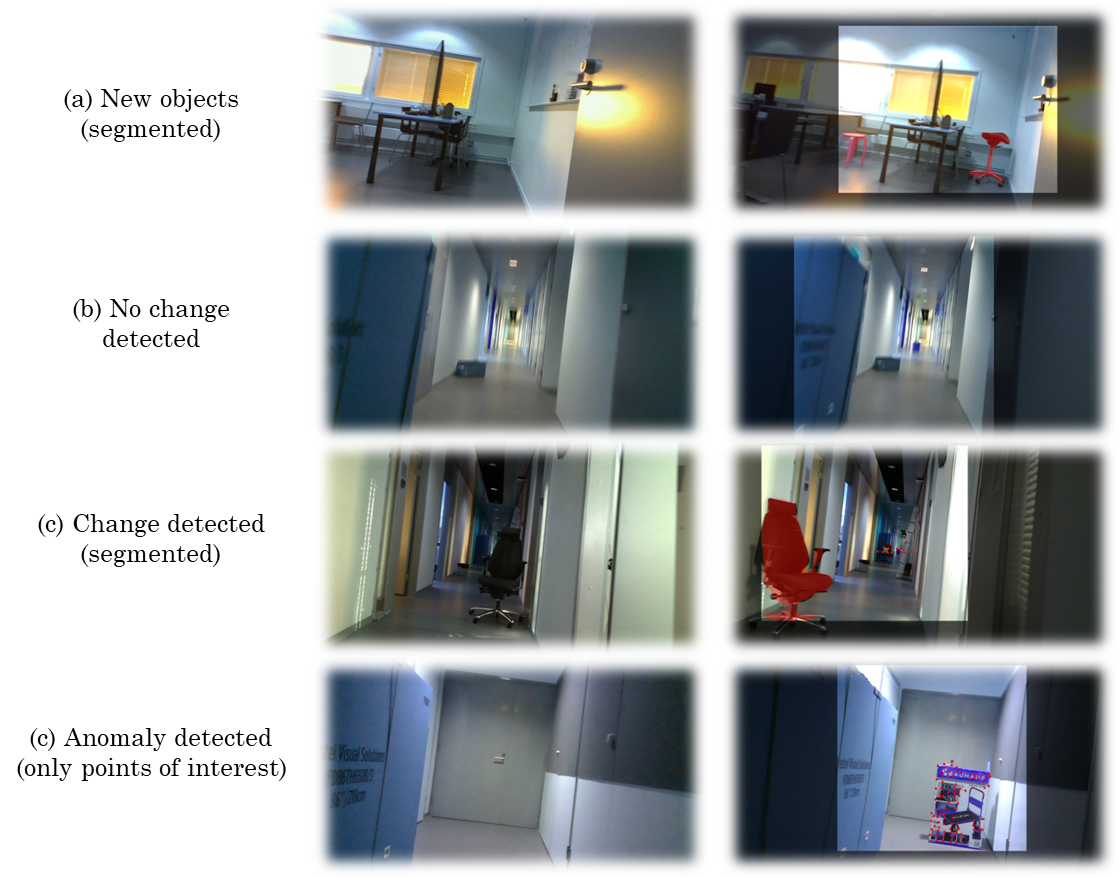}
    \caption{Selection of results from the method introduced in this work.}
    \label{fig:sample_results}
    \vspace{-1em}
\end{figure}

Machine learning methods have been extensively explored as a powerful tool for detecting anomalies. Based on reference information with labelled normal and abnormal data in the training process, deep anomaly detection models can be divided into three categories: supervised, semi-supervised, and unsupervised. These anomaly detection models often require a large number of training data from both normal and abnormal data to be effective in detecting anomalies~\cite{chalapathy2019deep}. The supervised models are trained for normal and abnormal images, as well as for limited foreign objects inside abnormal images. The unsupervised models are also trained for normal data as a single-class classification, then can classify other classes as the alien classes~\cite{perera2019learning}.

Some of the main limitations of existing works on visual anomaly detection tasks are that they have trained their models on specific environments and datasets~\cite{wang2022deep}, and the models for a large variety of datasets need to have a robust localization or a very similar viewpoint~\cite{avola2021low}. Also, works that are robust against environmental changes are trained for specific environments~\cite{slavic2021multilevel}.

% Vision-based anomaly detection and change detection in robotics has been implemented in various long-term autonomous operations such as inspection, monitoring, and surveillance systems~\cite{zaheer2021anomaly}. In such operations, robots may explore a completely unknown indoor and outdoor area without any previous knowledge of the environmental conditions. Therefore, the detection of byzantine data can be cast into the detection of tampered changes between two pairs of images. The availability of detecting anomalous data at the beginning of multi- and single-robot missions can be used as a baseline for comparison a posteriori. Within this context, we refer to anomalous data as both changes with respect to a reference dataset from the environment (e.g., moved objects) as well as the detection of foreign objects or other anomalies.

In this study, we focus on a vision-based anomaly and change detection system for finding abnormal data based on detecting foreign or changed objects at the pixel level in single-robot missions (see Fig.~\ref{fig:sample_results}). Anomalies or changes are both detected but not separately classified. We target the application domain of inspection robots. We first assume that the robot is able to navigate through its environment and records normal reference images at least once. We then provide a method for self-calibration of the model based on the reference data, and a system for comparing any future images with corresponding images captured from similar locations in the reference data to detect abnormal objects or changes. The focus of our analysis has been pixel-level change detection in the second robot's camera. The evaluation was conducted on pairs of images with from different cameras and with different conditions including variable viewpoints, distance to the anomaly, and environment lightning.

When robots move in different directions and follow different routes, there can be noticeable changes in the output images, such as viewpoint changes. To overcome the mentioned challenges, a method has been applied to extract and compare multi-scale features between the input image and the corresponding image in the trusted robot. We have employed SuperPoint~\cite{detone2018superpoint}, and SuperGlue~\cite{sarlin2020superglue} for feature extraction and feature matching process between pairs of images and calibrated them based on different environments and cameras. These features are robust to lighting, scale, and viewpoint changes. Then, in order to recognize the exact object as the anomaly in the image, the trained Mask-RCNN instance segmentation model has been used to segment anomalous parts based on the extracted features. Due to the limited number of object classes in each model and the fact that there are always untrained classes, in order to address the detection of unknown anomalies, DBSAN clustering method also is applied on unmatched interest points that are not segmented. Final anomalies are detected using both the segmentation and clustering processes.

% The core contributions of this work are the following:***

The rest of this manuscript is structured as follows. Section II discusses related work in the relevant anomaly and change detection literature. The background of the study is discussed in section III, and section IV provides an overview of the methodological approaches we used for this analysis. Results are presented in Section V, and Section VI concludes the work and points to future directions.
% \begin{figure}
%     \centering
%     \includegraphics[width=0.49\textwidth]{fig/cooperative_v23.pdf}
%     \caption{Cooperative localization approach based on UWB ranging measurements from multiple transceivers in different robots}
%     \label{fig:concept}
% \end{figure}

%%%%%%%%%%%%%%%%%%%%%%%%%%%%%%%%%%%%%%%%%%%%%%
%%                                          %%
%%              RELATED WORKS               %%
%%                                          %%
%%%%%%%%%%%%%%%%%%%%%%%%%%%%%%%%%%%%%%%%%%%%%%

\section{Related Work} \label{sec:related_work}

Recent studies have focused mostly on using image reconstruction approaches based on convolutional neural network models to detect visual anomalies in a certain class. Autoencoders and generative adversarial networks (GAN) are popular deep generative models. These models are trained on sufficient normal images of a single class to be able to reconstruct test images of the same class and detect the abnormal classes by comparing the loss scores~\cite{schlegl2017unsupervised,zavrtanik2021reconstruction}. In~\cite{wang2022deep} the authors applied a three-step procedure based on a deep generative model including an autoencoder and discriminator to detect abnormal images and foreign objects in rail images. A cognitive visual anomaly detection model consisting of an auto-encoder was utilized for industrial inspection robot to detect abnormal images with larger reconstruction errors than normal images in~\cite{li2020cognitive}.

For pixel-level anomaly detection, some studies used segmentation-based approaches. The authors in~\cite{di2021pixel} developed an anomaly detection method based on the comparison of features between the input and a generated photo-realistic image using a semantic segmentation model for road images. In~\cite{franklin2020anomaly,pustokhina2021automated} the Mask-RCNN segmentation model was used to detect anomaly events and instances for video surveillance systems. However, when anomaly is a missed object or an unknown object, it cannot be detected or segmented by the segmentation and detection methods.

Change detection is also widely used in several fields, including aerial image change detection~\cite{liu2020deep}, urban change monitoring~\cite{zhou2021novel}, and anomaly detection~\cite{wang2022anodfdnet} by comparing images of the same area. Wang et al. in~\cite{wang2021transcd} introduced a scene change detection model using the siamese vision transformer and CNN model to generate corresponding pixel-wise change maps. Similar models have been designed to be robust to outdoor conditions, including illumination, scaling, and viewpoint changes~\cite{guo2018learning}. ChangeNet~\cite{varghese2018changenet}, and CSCDNet~\cite{sakurada2020weakly} which are deep siamese networks for detecting changes between pairs of images, have been used for different datasets and applications~\cite{park2021changesim}.

In a recent study~\cite{takeda2022domain}, the authors presented a visual change detection method for robotics applications. They applied an attention mask to the intermediate layer of a siamese CNN to detect small changes in pairs of images. They evaluated their method by comparing reference images to live images with slight variations in viewpoint in indoor environment. 
A number of studies have used feature detection methods to detect anomalies in pair images~\cite{lu2021anomaly}. 

There are a number of well-known classical feature detectors in the computer vision area, including SIFT~\cite{lowe2004distinctive}, SURF~\cite{bay2006surf}, and ORB~\cite{rublee2011orb}, as well as deep neural network-based methods, such as SuperPoint. An anomaly detection model is described in~\cite{zhang2016abnormal} using ORB features and sliding windows. In a relevant study~\cite{zhang2022image}, SIFT feature extractor along with polar cosine transform (PCT) have been incorporated to detect tempered pixels in images for internet of things (IoT) security. A fully-convolutional neural network architecture called SuperPoint which is a self-supervised feature point detector and descriptor was presented in~\cite{detone2018superpoint}. SuperGlue~\cite{sarlin2020superglue} as a keypoints matching technique showed better performance in conjunction with SuperPoint to match two sets of extracted features and corresponding descriptors. The results showed that their proposed method tends to generate a larger number of correct matches that broadly cover the image compared to other traditional methods. Both SuperPoint and SuperGlue have different confidence parameters such as keypoint detectors and matching confidence thresholds which have a major effect on their performance and the final results.

%%%%%%%%%%%%%%%%%%%%%%%%%%%%%%%%%%%%%%%%%%%%%%
%%                                          %%
%%        PROBLEM DEFINITION                %%
%%                                          %%
%%%%%%%%%%%%%%%%%%%%%%%%%%%%%%%%%%%%%%%%%%%%%%

\begin{figure*}
    \centering
    \includegraphics[width=\textwidth]{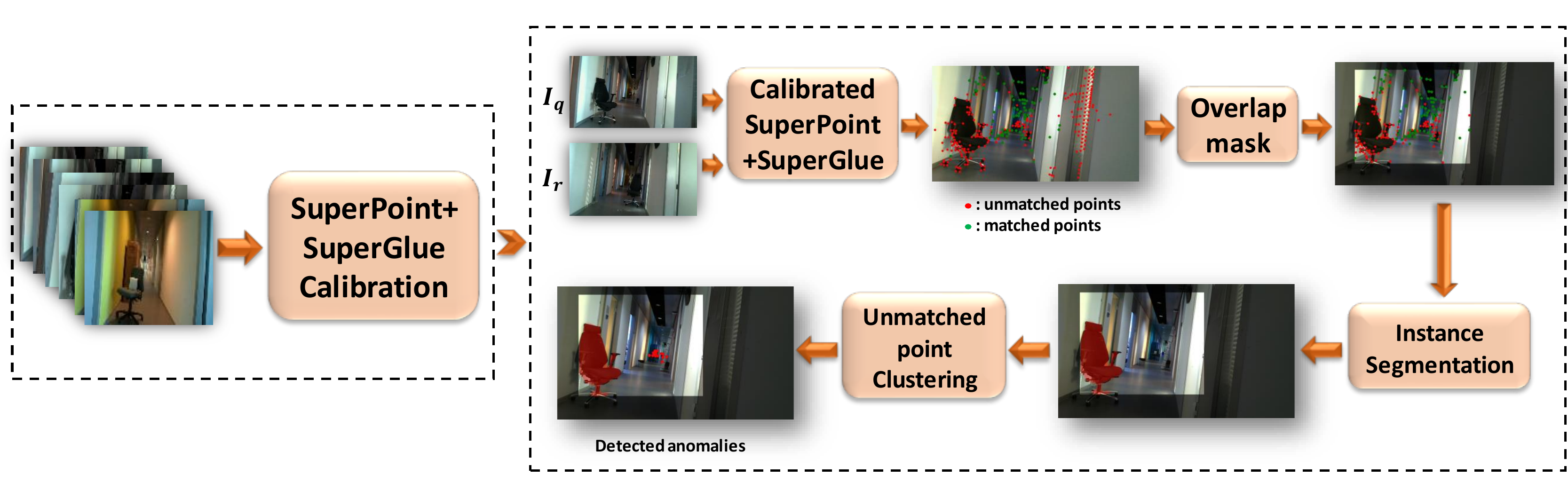}
    \caption{Architectural diagram of the proposed framework.}
    \label{fig:diagram}
\end{figure*}

\section{Background}

Through this section, we introduce the key neural network architectures that are employed in our work.

\subsection{SuperPoint}

SuperPoint is a self-supervised fully convolutional neural network for extracting interest points and their descriptors. Its architecture starts with pre-training a base detector called MagicPoint on synthetically generated dataset that includes simple shapes such as rectangles, stars, and cubes to extract interest points. Detecting corners of simple geometric shapes using the initial base detector performs well even with added noise~\cite{detone2018superpoint}. However, the base detector misses many important interest point in real images. For better generalization, it is combined with homographic adaptation to provide more training samples from each image in the MS-COCO dataset. Using this approach, pseudo-ground truth interest points are generated for each image. The final network has a shared VGG-based encoder and two decoders. The first decoders is for point detection, while the second one is for point description. There are multiple shared parameters between both. Compared to traditional feature point extractors such as LIFT, ORB, or SIFT, the Superpoint model has recently achieved superior %impressive 
results in several research projects.

\subsection{SuperGlue}

SuperGlue is a feature point matching method that takes key points and their descriptors in image pairs and matches them with corresponding points using a graph neural network. It has been shown to produce better results in combination with the SuperPoint-generated features. In short, the algorithm is composed of two parts, an attentional graph neural network, and an optimal matching layer. A differentiable Sinkhorn algorithm is used to efficiently pair matchable points and reject non-matchable points at the final step~\cite{sarlin2020superglue}. SuperPoint and Superglue models require predefined thresholds for different parameters such as sinkhorn iterations, non-maximum suppression (NMS), and keypoints and matching thresholds. These parameters affect the performance of the networks, and therefore are often tailored to the nature of the data being fed to the models. In this work we introduce a self-calibration approach given some data samples. %which are depends on various data.

\subsection{Instance segmentation}

Object segmentation is a process in which a specific class is assigned to pixel values of an image, and it is broadly divided into two types: semantic segmentation and instance segmentation. Semantic segmentation results in all pixels of the same class with the same value. Instance segmentation methods, instead, identify multiple objects of a single class as distinct instances of interest in an image, and only label pixels of classified objects. Recently, several approaches for instance segmentation have been proposed, with the most popular technique still relying on Mask R-CNN, a two-stage detection and segmentation approach~\cite{he2017mask}, robust to different image types~\cite{xianjia2022analyzing}. In this approach, and using an object detection model, bounding boxes are first predicted for all instances. Then, regions-of-interest (ROIs) are cropped for each instance and all ROIs are then fed to a fully convolutional network for foreground and background segmentation.

%%%%%%%%%%%%%%%%%%%%%%%%%%%%%%%%%%%%%%%%%%%%%%
%%                                          %%
%%              METHODOLOGY                 %%
%%                                          %%
%%%%%%%%%%%%%%%%%%%%%%%%%%%%%%%%%%%%%%%%%%%%%%

\section{Methodology}

In this section, we introduce a calibration procedure for the feature extraction algorithm, along with the segmentation and clustering of interest points. A diagram of the proposed process is outlined in Fig.~\ref{fig:diagram}.

For detecting both changes and foreign%byzantine 
objects in an environment, two sets of images are processed based on the SuperPoint and the SuperGlue techniques. We calibrated the models using a single-shot procedure since the number of matched and non-matched points varies depending on different thresholds. In this study, we have used the SuperGlue GitHub repository\footnote{\url{https://github.com/magicleap/SuperGluePretrainedNetwork}} to extract and match keypoints of base and test images. 

The main parameter to adjust is the SuperGlue match threshold ($\Delta$) that directly affects the number of matched interest points. We have also analyzed the impact of other parameters but found any significant performance changes. A small $\Delta$ close to $0$ (e.g., the default value $\Delta=0.2$ used in the SuperGlue paper) results in wrong common overlap areas in image pairs. A large threshold (close to $1$) , on the other hand, leads to incorrect anomaly detection. We change the amount of matching threshold from $\Delta=0$ to $\Delta=0.9$ with steps of 0.1. This parameter must be calibrated according to different environmental conditions and camera types. To this end, we capture a few image pairs with a horizontal shift %were captured 
with three different cameras. The images are obtained under various conditions, with changes in lightning and environment structure or background. Due to the linear shift between image pairs, we  hypothesize that appropriate thresholds should yield a low standard deviation of the distribution of distances between matched keypoint pairs.  Let \(N\) be the number of image pairs and \(mp_n\) be the number of matched interest points in image pair number \(n\). The coefficient of variation (\(cv\)) of the image pairs can be computed following Eq.~\eqref{eq:cv_n}:

\begin{equation}
    \label{eq:cv_n}
    cv_n = \frac{\sigma_n}{\mu_n}   \quad n = 0,...,N
\end{equation}

where \(\sigma_n\) and \(\mu_n\) are the standard deviation and the average distance between matched keypoints, respectively. If the number of matched interest points is close to $0$, it will result in a smaller amount of \(cv_n\) so, for the ideal threshold values with a maximum number of interest points, each \(cv_n\) is divided by the number of matched interest points. The average of this value for \(N\) image pairs is calculated as given by Eq.~\eqref{eq:m_cv}:

\begin{equation}
    \label{eq:m_cv}
    cv = \frac{\sum_{n=1}^{N} {\frac{cv_n}{mp_n}}}{N} \quad n = 0,...,N
\end{equation}

The deployment, calibration and anomaly detection process then proceeds as follows:

\textbf{1. Baseline data:} once a robot is deployed, we first gather a sample set of images of the new environment, which can be as small as a single pair of images. We then measured \(cv\) for different matching thresholds. Low thresholds result in more true positive, but also more false positives. The best value is a balance point between maximum number of keypoints and minimum amount of \(cv\). We use the Kneedle algorithm to find a balance point on the curve of the mean of the coefficient of variation per matching threshold. This algorithm selects the so-called knee point. This point is defined as the furthest away from a line defined by the higher and lower points with maximum curvature~\cite{satopaa2011finding}. 

\textbf{2. Clean run:} before starting the anomaly detection process, the robot performs a so-called \textit{clean run} of the operational environment, which we assume to be unchanged at this state. We use this set of images as the reference for comparing any future missions, and detect anomalies

\textbf{3. Image pair matching:} Using onboard localization methods or other source of positioning information, relevant image pairs of two paths are selected based on both position and orientation of the robot. These images are then fed to the SuperPoint and SuperGlue algorithms with ideal threshold for matching. The positions and orientations of these image pairs are similar, but the frames aren't exactly the same, which would lead to false anomaly detections. In order to determine and process the overlap area in the query image compared to the reference image, a mask is created using the maximum and minimum matched interest points in x and y coordinates. 

\textbf{4. Instance segmentation:} In the next step, the Mask-RCNN model is used to segment different instances in query image and the result is combined with extracted keypoints. The pre-trained Mask-RCNN-X101-FPN model used in this study is from the Facebook AI Research library called Detectron2, which implements most of the state-of-the-art object detection and segmentation algorithms~\cite{wu2019detectron2}. Since this model has been trained on a huge number of COCO images, it can segment most of the probable object classes.

\textbf{5. Anomalous object identification:} Using segmented masks, extracted unmatched keypoints can be clustered and mapped to a specific anomalous object. Each segmented object is analyzed based on the number of matched and not matched keypoints. If the number of unmatched keypoints in an object is more, then it is considered as an anomaly. However, there are still unknown objects that cannot be segmented using any of the existing instance segmentation models, which have been trained on a wide range of classes. 

In order to detect all anomalies, regardless of their classes, we apply a density-based clustering method to the unmatched interest points. At this point, we delete segmented unmatched points, and all points that remain are grouped using density-based spatial clustering of applications with noise (DBSCAN)~\cite{ester1996density}. The DBSCAN algorithm estimates the minimum density level based on the number of neighborhood points, minPts, within a certain distance threshold, Eps. An anomaly refers to a group of unmatched interest points within this distance threshold with more than minPts neighbors. A summary of the described process can be found in Algorithm~\ref{alg:anomaly_detection}.

\begin{algorithm}[t]
    \small
	\caption{anomaly detection in image sequences}
	\label{alg:anomaly_detection}
	\KwIn{\\
	    Reference images: \textit{r}\\
	    Reference positions: \textit{r\_p}\\
	    Query images with anomaly: \textit{q}\\
	    Query positions: \textit{q\_p}\\
	    Range of matching and keypoint thresholds: m, k \\
	    Range of distance and orientation thresholds: d, o \\ 
	}
	\textbf{Calibration:}\\
	\hspace{1em}SuperGlue\_calibration();\\
	\hspace{1em}DBSCAN\_clusterring();\\
	\textbf{Geo\_information:}\\
	\hspace{1em}mapping();\\
	\hspace{1em}Geo\_relevant\_images();\\
    \For{r\_image and q\_image } {
        $calibrated\_SuperPoint\_and\_SuperGlue()$;\\
        $Overlap\_mask()$;\\
            \ForEach{overlapped\_query} {
                $Instance\_segmentation()$;\\
                \ForEach{instance\_mask} {
                \hspace{1em}n\_len = length(segmented\_unmatched);\\
                \hspace{1em}m\_len = length(segmented\_matched);\\
                \hspace{1em}\If{n\_len - m\_len $\geq$ 1} {
                    \hspace{1em}overlapped\_unmatched.remove(\\\hspace{1em}segmented\_unmatched);\\
                    \hspace{1em}anomaly $\leftarrow$ instance\_mask;\\
                    \hspace{1em}anomaly\_class $\leftarrow$ instance\_class;\\
                    
                }                
            }
            \hspace{1em}DBSCAN\_clusterring(overlapped\_unmatched);\\
        }
    }
\end{algorithm}

%%%%%%%%%%%%%%%%%%%%%%%%%%%%%%%%%%%%%%%%%%%%%%
%%                                          %%
%%              EXPERIMENTS                 %%
%%                                          %%
%%%%%%%%%%%%%%%%%%%%%%%%%%%%%%%%%%%%%%%%%%%%%%

\begin{figure}[t]
    \centering
    \includegraphics[width=.45\textwidth]{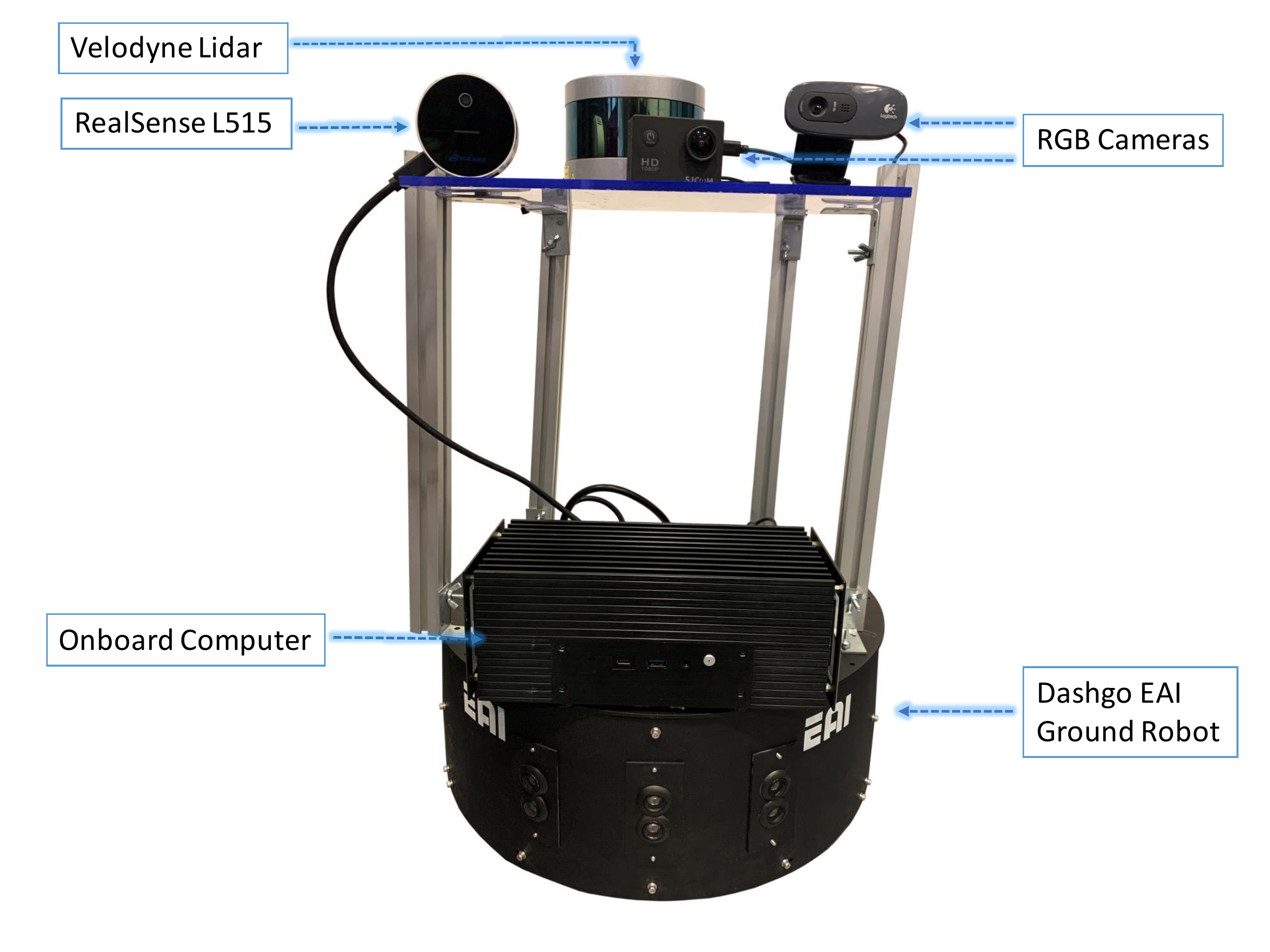}
     \caption{Equipment utilized for data collection. The three cameras utilized in the setup have different field of view, with one of them having a wide lens and only the L515 has a global shutter. The cameras are situated in three different vertical planes to reduce the overlap between the images and increase simulate additional orientation drift for the experiments.}
    \label{fig:platform}
\end{figure}

\section{Experimental Results}

%For the experiment procedure explained above, 

This section covers experiments that we carried out with a ground robot to demonstrate the effectiveness of the proposed anomaly detection and model calibration methods.

\textit{Hardware.} We mounted two commercial RGB different cameras and a RealSense L515 lidar camera on a Dashgo ground robot to determine the optimum thresholds for the test environment. The robot is also equipped with a Velodyne lidar to obtain the position and orientation information using existing lidar odometry and mapping methods~\cite{li2020multi, li2022multi}. The platform is shown in~\ref{fig:platform}. 

\textit{Software.} The system has been implemented using ROS Melodic under Ubuntu 18.04. The robot is commanded to explore an environment and record the reference RGB images, and lidar point cloud data. Later, the robot is commanded to scan the same place in a recurrent manner to detect anomalous objects or changes. We assume that there might be slight deviations between the paths followed by the robot in consecutive iterations. Figure~\ref{fig:map} shows the robot's trajectory during the collection of the reference data and the anomaly detection phases and the green points inside the map show different inserted anomaly objects. 

\input{tex/map.tex}

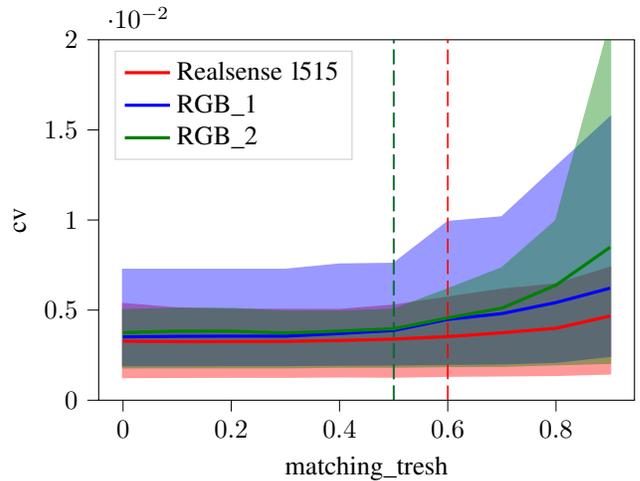
\begin{figure}
    \centering
    \begin{tikzpicture}
    
    \definecolor{darkgray176}{RGB}{176,176,176}
    \definecolor{green}{RGB}{0,128,0}
    \definecolor{lightgray204}{RGB}{204,204,204}
    
    \begin{axis}[
        width=.48\textwidth,
        height=.27\textheight,
    legend cell align={left},
    legend style={
      fill opacity=0.8,
      draw opacity=1,
      text opacity=1,
      at={(0.03,0.97)},
      anchor=north west,
      draw=lightgray204
    },
    tick align=outside,
    tick pos=left,
    x grid style={darkgray176},
    xlabel={matching\_tresh},
    xmin=-0.045, xmax=0.945,
    xtick style={color=black},
    y grid style={darkgray176},
    ylabel={cv},
    ymin=0, ymax=0.02,
    ytick style={color=black}
    ]
    \path [draw=red, fill=red, opacity=0.4, thick]
    (axis cs:0,0.0053419346889753)
    --(axis cs:0,0.00127951736501653)
    --(axis cs:0.1,0.00129006388815732)
    --(axis cs:0.2,0.00129844618795571)
    --(axis cs:0.3,0.00129758418151186)
    --(axis cs:0.4,0.00131833188387812)
    --(axis cs:0.5,0.00130401214706564)
    --(axis cs:0.6,0.00135355507919544)
    --(axis cs:0.7,0.00137373232147069)
    --(axis cs:0.8,0.00139686346529015)
    --(axis cs:0.9,0.00148074009362806)
    --(axis cs:0.9,0.00734897679889325)
    --(axis cs:0.9,0.00734897679889325)
    --(axis cs:0.8,0.00642730956687067)
    --(axis cs:0.7,0.00613555714890764)
    --(axis cs:0.6,0.00569802332835592)
    --(axis cs:0.5,0.00525131892590296)
    --(axis cs:0.4,0.00501089082287915)
    --(axis cs:0.3,0.0050186525131094)
    --(axis cs:0.2,0.0050186525131094)
    --(axis cs:0.1,0.00509657723563058)
    --(axis cs:0,0.0053419346889753)
    --cycle;
    
    \path [draw=blue, fill=blue, opacity=0.4, thick]
    (axis cs:0,0.00723881400548495)
    --(axis cs:0,0.00193899504418643)
    --(axis cs:0.1,0.00193899504418643)
    --(axis cs:0.2,0.00194950348941992)
    --(axis cs:0.3,0.00194950348941992)
    --(axis cs:0.4,0.0019649216349091)
    --(axis cs:0.5,0.00197973989305042)
    --(axis cs:0.6,0.00201226181263546)
    --(axis cs:0.7,0.00204174468914668)
    --(axis cs:0.8,0.00212804019918018)
    --(axis cs:0.9,0.00245650055820991)
    --(axis cs:0.9,0.0156835132631762)
    --(axis cs:0.9,0.0156835132631762)
    --(axis cs:0.8,0.0129319658646217)
    --(axis cs:0.7,0.0101514225420745)
    --(axis cs:0.6,0.00989254277486068)
    --(axis cs:0.5,0.00757057099573074)
    --(axis cs:0.4,0.00753139844164252)
    --(axis cs:0.3,0.00723881400548495)
    --(axis cs:0.2,0.00723881400548495)
    --(axis cs:0.1,0.00723881400548495)
    --(axis cs:0,0.00723881400548495)
    --cycle;
    
    \path [draw=green, fill=green, opacity=0.4, thick]
    (axis cs:0,0.00500203338829247)
    --(axis cs:0,0.00181843489408493)
    --(axis cs:0.1,0.00182265690323704)
    --(axis cs:0.2,0.00182265690323704)
    --(axis cs:0.3,0.00182265690323704)
    --(axis cs:0.4,0.00185695627244447)
    --(axis cs:0.5,0.00185695627244447)
    --(axis cs:0.6,0.00189129524170213)
    --(axis cs:0.7,0.00190183227388268)
    --(axis cs:0.8,0.00199224216710702)
    --(axis cs:0.9,0.00206944554351097)
    --(axis cs:0.9,0.0208600185535572)
    --(axis cs:0.9,0.0208600185535572)
    --(axis cs:0.8,0.00994895725715451)
    --(axis cs:0.7,0.00732320026745872)
    --(axis cs:0.6,0.00615278904030963)
    --(axis cs:0.5,0.00502716129718546)
    --(axis cs:0.4,0.00491863161652953)
    --(axis cs:0.3,0.00491863161652953)
    --(axis cs:0.2,0.00507264584302902)
    --(axis cs:0.1,0.00507264584302902)
    --(axis cs:0,0.00500203338829247)
    --cycle;
    
    \path [draw=red, semithick, dash pattern=on 5.55pt off 2.4pt]
    (axis cs:0.6,0)
    --(axis cs:0.6,1);
    
    \path [draw=blue, semithick, dash pattern=on 5.55pt off 2.4pt]
    (axis cs:0.5,0)
    --(axis cs:0.5,1);
    
    \path [draw=green, semithick, dash pattern=on 5.55pt off 2.4pt]
    (axis cs:0.5,0)
    --(axis cs:0.5,1);
    
    \addplot [very thick, red]
    table {%
    0 0.00327517407551983
    0.1 0.00325033262370047
    0.2 0.00325517626842982
    0.3 0.00326472915063283
    0.4 0.00331264057668528
    0.5 0.00338967829808426
    0.6 0.00352776074361386
    0.7 0.00374163534842254
    0.8 0.00399053854531363
    0.9 0.00467232784095282
    };
    \addlegendentry{Realsense l515}
    \addplot [very thick, blue]
    table {%
    0 0.00351391206298132
    0.1 0.00354361888576765
    0.2 0.00355410914752297
    0.3 0.00355294954649995
    0.4 0.00369996215973579
    0.5 0.00387023610650721
    0.6 0.00448170808788852
    0.7 0.004806061345135
    0.8 0.00541936730004712
    0.9 0.00621786625701838
    };
    \addlegendentry{RGB\_1}
    \addplot [very thick, green]
    table {%
    0 0.00375158730203256
    0.1 0.00382898306979017
    0.2 0.00383194967357647
    0.3 0.00373130992864695
    0.4 0.00383722079794274
    0.5 0.00396350548442644
    0.6 0.0045654557768739
    0.7 0.00511034621820423
    0.8 0.00638459979932031
    0.9 0.00849033342359757
    };
    \addlegendentry{RGB\_2}
    \end{axis}
    
    \end{tikzpicture}
    
\caption{Average thresholds for three different cameras for SuperPoint+SuperGlue calibration.}
\label{fig:thresh}
\end{figure}

\textit{Calibration.} Figure~\ref{fig:thresh} shows the optimum matching thresholds to calibrate SuperPoint + SuperGlue model. A few image pairs with linear shift have been recorded using three different cameras in the same environment. For each camera, average threshold values were calculated along with the minimum and maximum ranges of coefficients of variation (shaded areas) in pair images. For the RealSense L515 camera, the best value is $\Delta=0.6$, and for the other two RGB cameras, the best value is $\Delta=0.5$. In both cases the selected value differs significantly from the default value of $\Delta=0.2$. 
    % This file was created with tikzplotlib v0.10.1.

It is worth noting that the DBSCAN algorithm performance depends heavily on the choice of minPts and distance threshold parameters. We have used the k-nearest neighbor's approach mentioned in the referenced paper to find the optimal distance threshold. In Fig.~\ref{fig:dbscan}, the average Eps has been determined with a minimum of 5 neighborhood points for some random images with anomalies. Small anomalous objects cannot be clustered by selecting a larger number of minPts.

\begin{figure*}
    \centering
    \includegraphics[width=\textwidth]{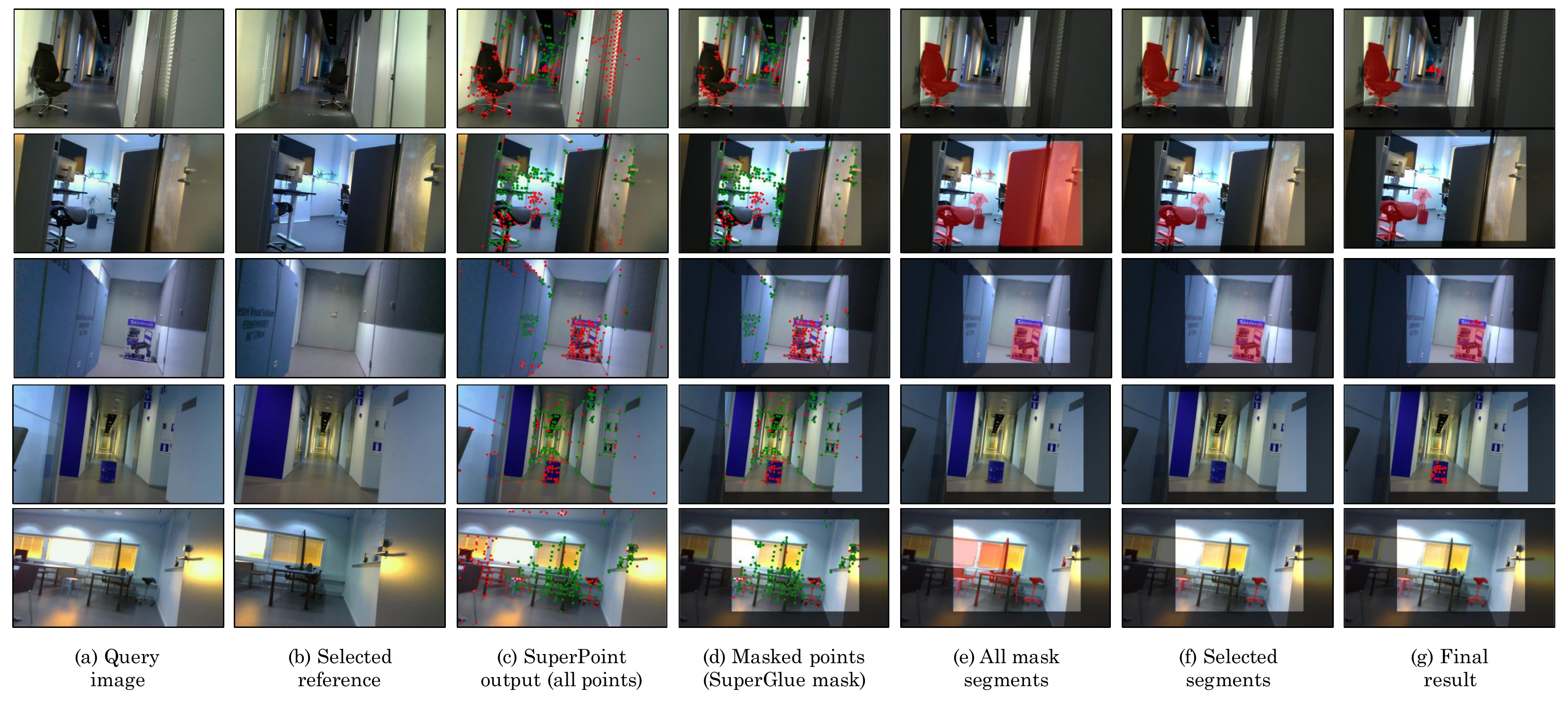}
    \caption{Selection of images in the experiment featuring different environments and types of anomaly.}
    \label{fig:final_results}
\end{figure*}

\input{tex/dbscan.tex}

\input{tex/accuracy.tex}

\textit{Results.} When the feature detection model has been calibrated based on the optimal matching threshold, RGB images of both reference and query data are recorded using the realsense l515 camera to evaluate the proposed approach. Figure~\ref{fig:accuracy} shows that the proposed approach achieved 72\% accuracy in detecting anomalies in more than 270 image sequences over time. In some cases, for unknown small and textureless anomalous objects that cannot also be detected and segmented by the segmentation model, the proposed approach shows low accuracy for time ranges between 40\,s and 90\,s (see Fig.~\ref{fig:accuracy}). In this figure, the performance of the proposed approach has also been analyzed for different matching thresholds. The calibrated feature extraction model with the matching threshold of $\Delta=0.6$ achieved higher accuracy, demonstrating the effectiveness of the proposed autocalibration method. It is worth noting that the reference images are selected from the reference dataset based on the robot's position and orientation. This is a limitation as a localization error might render the method unusable. However, our method is robust to viewpoint and therefore only major errors would have an impact. In those cases, still, our method would potentially detect anomalies across the entire image, so the results can provide feedback to the localization process or robot operator indicating that there might be a large error. 

Figure~\ref{fig:final_results} shows a selection of the final results of the proposed anomaly and change detection workflow. The figure includes the main steps taken in the process: (a) the query image, (b) the selected reference image form the \textit{clean} dataset based on position and orientation, (c) all the matched and unmatched points returned by SuperPoint, (d) the points that fall under the SuperGlue-generated mask, (e) the segments detected in the masked area with Mask-RCNN, (f) the selected segments where enough unmatched points are present, and (g) the segments detected as anomalies and unsegmented by clustered unmatched points that also represent potential anomalies or changes. We can see that our method works for various environmental conditions, and is able to detect anomalies and changes as segmented objects or sets of points of interest. The segmentation approach enables the detection of anomalies in largely feature-\textit{less} objects (e.g., large flat objects such as TVs) where a clustering approach does not work. On the other hand, the clustering of points in non-segmented image areas aids in detecting changes or anomalies regarding objects that the segmentation network is not able to classify. Overall, these two approaches together with the self-calibration show a promising way of general-purpose anomaly and change detection for mobile robots.

\section{Conclusion}\label{sec:conclusion}

In this paper, a visual anomaly and change detection approach for detecting pixel-level anomalies in image pairs has been presented. Using a single-shot method we have calibrated SuperPoint and SuperGlue feature detection models in order to find the proper unmatched interest points. For detecting anomalous regions and objects in the query image, a general instance segmentation model has been applied to unmatched points, as well as the DBSCAN method clustered remain points that do not belong to any object. According to the experimental results, it has been shown that this framework yields promising accuracy for a general and unknown environment and has no need for training on specific data.

In future work, we will analyze the performance across more different environments. We will also work on deploying this anomaly detection framework in a distributed manner in order to identify potentially byzantine robots within a larger multi-robot system, without a reference dataset.

%%%%%%%%%%%%%%%%%%%%%%%%%%%%%%%%%%%%%%%%%%%%%%
%%                                          %%
%%            ACKNOWLEDGMENT                %%
%%                                          %%
%%%%%%%%%%%%%%%%%%%%%%%%%%%%%%%%%%%%%%%%%%%%%%

\section*{Acknowledgment}

This research work is supported by the R3Swarms project funded by the Secure Systems Research Center (SSRC), Technology Innovation Institute (TII).

%%%%%%%%%%%%%%%%%%%%%%%%%%%%%%%%%%%%%%%%%%%%%%
%%                                          %%
%%              BIBLIOGRAPHY                %%
%%                                          %%
%%%%%%%%%%%%%%%%%%%%%%%%%%%%%%%%%%%%%%%%%%%%%%
% \newpage
% \nocite{*}
\bibliographystyle{abbrv}
\bibliography{bibliography}

\begin{thebibliography}{10}

\bibitem{akcay2018ganomaly}
S.~Akcay, A.~Atapour-Abarghouei, and T.~P. Breckon.
\newblock Ganomaly: Semi-supervised anomaly detection via adversarial training.
\newblock In {\em Asian conference on computer vision}. Springer, 2018.

\bibitem{avola2021low}
D.~Avola, L.~Cinque, A.~Di~Mambro, A.~Diko, A.~Fagioli, G.~L. Foresti, M.~R.
  Marini, A.~Mecca, and D.~Pannone.
\newblock Low-altitude aerial video surveillance via one-class svm anomaly
  detection from textural features in uav images.
\newblock {\em Information}, 13(1), 2021.

\bibitem{bay2006surf}
H.~Bay, T.~Tuytelaars, and L.~V. Gool.
\newblock Surf: Speeded up robust features.
\newblock In {\em European conference on computer vision}. Springer, 2006.

\bibitem{chalapathy2019deep}
R.~Chalapathy and S.~Chawla.
\newblock Deep learning for anomaly detection: A survey.
\newblock {\em arXiv preprint arXiv:1901.03407}, 2019.

\bibitem{detone2018superpoint}
D.~DeTone, T.~Malisiewicz, and A.~Rabinovich.
\newblock Superpoint: Self-supervised interest point detection and description.
\newblock In {\em IEEE CVPR workshops}, 2018.

\bibitem{di2021pixel}
G.~Di~Biase, H.~Blum, R.~Siegwart, and C.~Cadena.
\newblock Pixel-wise anomaly detection in complex driving scenes.
\newblock In {\em Proceedings of the IEEE/CVF conference on computer vision and
  pattern recognition}, 2021.

\bibitem{ester1996density}
M.~Ester, H.-P. Kriegel, J.~Sander, X.~Xu, et~al.
\newblock A density-based algorithm for discovering clusters in large spatial
  databases with noise.
\newblock In {\em kdd}, volume~96, 1996.

\bibitem{fernando2020deep}
T.~Fernando, H.~Gammulle, S.~Denman, S.~Sridharan, and C.~Fookes.
\newblock Deep learning for medical anomaly detection--a survey.
\newblock {\em arXiv preprint arXiv:2012.02364}, 2020.

\bibitem{franklin2020anomaly}
R.~J. Franklin, V.~Dabbagol, et~al.
\newblock Anomaly detection in videos for video surveillance applications using
  neural networks.
\newblock In {\em 2020 Fourth International Conference on Inventive Systems and
  Control (ICISC)}. IEEE, 2020.

\bibitem{guo2018learning}
E.~Guo, X.~Fu, J.~Zhu, M.~Deng, Y.~Liu, Q.~Zhu, and H.~Li.
\newblock Learning to measure change: Fully convolutional siamese metric
  networks for scene change detection.
\newblock {\em arXiv preprint arXiv:1810.09111}, 2018.

\bibitem{he2017mask}
K.~He, G.~Gkioxari, P.~Doll{\'a}r, and R.~Girshick.
\newblock Mask r-cnn.
\newblock In {\em Proceedings of the IEEE international conference on computer
  vision}, 2017.

\bibitem{kamat2020anomaly}
P.~Kamat and R.~Sugandhi.
\newblock Anomaly detection for predictive maintenance in industry 4.0-a
  survey.
\newblock In {\em E3S web of conferences}, volume 170. EDP Sciences, 2020.

\bibitem{li2020cognitive}
J.~Li, X.~Xu, L.~Gao, Z.~Wang, and J.~Shao.
\newblock Cognitive visual anomaly detection with constrained latent
  representations for industrial inspection robot.
\newblock {\em Applied Soft Computing}, 95, 2020.

\bibitem{li2020multi}
Q.~Li, J.~{Pe{\~n}a Queralta}, T.~N. Gia, Z.~Zou, and T.~Westerlund.
\newblock Multi-sensor fusion for navigation and mapping in autonomous
  vehicles: Accurate localization in urban environments.
\newblock {\em Unmanned Systems}, 8(03):229--237, 2020.

\bibitem{li2022multi}
Q.~Li, X.~Yu, J.~{Pe{\~n}a Queralta}, and T.~Westerlund.
\newblock Multi-modal lidar dataset for benchmarking general-purpose
  localization and mapping algorithms.
\newblock {\em arXiv preprint arXiv:2203.03454}, 2022.

\bibitem{liu2020deep}
R.~Liu, D.~Jiang, L.~Zhang, and Z.~Zhang.
\newblock Deep depthwise separable convolutional network for change detection
  in optical aerial images.
\newblock {\em IEEE Journal of Selected Topics in Applied Earth Observations
  and Remote Sensing}, 13, 2020.

\bibitem{lowe2004distinctive}
D.~G. Lowe.
\newblock Distinctive image features from scale-invariant keypoints.
\newblock {\em International journal of computer vision}, 60(2), 2004.

\bibitem{lu2021anomaly}
D.~Lu, X.~Liao, F.~Xu, and J.~Bai.
\newblock Anomaly detection method for substation equipment based on feature
  matching and multi-semantic classification.
\newblock In {\em 2021 6th Asia Conference on Power and Electrical Engineering
  (ACPEE)}. IEEE, 2021.

\bibitem{park2021changesim}
J.-M. Park, J.-H. Jang, S.-M. Yoo, S.-K. Lee, U.-H. Kim, and J.-H. Kim.
\newblock Changesim: Towards end-to-end online scene change detection in
  industrial indoor environments.
\newblock In {\em 2021 IEEE/RSJ International Conference on Intelligent Robots
  and Systems (IROS)}. IEEE, 2021.

\bibitem{perera2019learning}
P.~Perera and V.~M. Patel.
\newblock Learning deep features for one-class classification.
\newblock {\em IEEE Transactions on Image Processing}, 28(11), 2019.

\bibitem{pustokhina2021automated}
I.~V. Pustokhina, D.~A. Pustokhin, T.~Vaiyapuri, D.~Gupta, S.~Kumar, and
  K.~Shankar.
\newblock An automated deep learning based anomaly detection in pedestrian
  walkways for vulnerable road users safety.
\newblock {\em Safety science}, 142, 2021.

\bibitem{rublee2011orb}
E.~Rublee, V.~Rabaud, K.~Konolige, and G.~Bradski.
\newblock Orb: An efficient alternative to sift or surf.
\newblock In {\em 2011 International conference on computer vision}. Ieee,
  2011.

\bibitem{sakurada2020weakly}
K.~Sakurada, M.~Shibuya, and W.~Wang.
\newblock Weakly supervised silhouette-based semantic scene change detection.
\newblock In {\em 2020 IEEE International conference on robotics and automation
  (ICRA)}. IEEE, 2020.

\bibitem{sarlin2020superglue}
P.-E. Sarlin, D.~DeTone, T.~Malisiewicz, and A.~Rabinovich.
\newblock {SuperGlue}: Learning feature matching with graph neural networks.
\newblock In {\em CVPR}, 2020.

\bibitem{satopaa2011finding}
V.~Satopaa, J.~Albrecht, D.~Irwin, and B.~Raghavan.
\newblock Finding a" kneedle" in a haystack: Detecting knee points in system
  behavior.
\newblock In {\em 31st ICDCS Workshops}. IEEE, 2011.

\bibitem{schlegl2017unsupervised}
T.~Schlegl, P.~Seeb{\"o}ck, S.~M. Waldstein, U.~Schmidt-Erfurth, and G.~Langs.
\newblock Unsupervised anomaly detection with generative adversarial networks
  to guide marker discovery.
\newblock In {\em International conference on information processing in medical
  imaging}. Springer, 2017.

\bibitem{slavic2021multilevel}
G.~Slavic, M.~Baydoun, D.~Campo, L.~Marcenaro, and C.~Regazzoni.
\newblock Multilevel anomaly detection through variational autoencoders and
  bayesian models for self-aware embodied agents.
\newblock {\em IEEE Transactions on Multimedia}, 2021.

\bibitem{takeda2022domain}
K.~Takeda, K.~Tanaka, and Y.~Nakamura.
\newblock Domain invariant siamese attention mask for small object change
  detection via everyday indoor robot navigation.
\newblock {\em arXiv preprint arXiv:2203.15362}, 2022.

\bibitem{varghese2018changenet}
A.~Varghese, J.~Gubbi, A.~Ramaswamy, and P.~Balamuralidhar.
\newblock Changenet: A deep learning architecture for visual change detection.
\newblock In {\em Proceedings of the European Conference on Computer Vision
  (ECCV) Workshops}, 2018.

\bibitem{wang2022deep}
T.~Wang, Z.~Zhang, and K.-L. Tsui.
\newblock A deep generative approach for rail foreign object detections via
  semi-supervised learning.
\newblock {\em IEEE Transactions on Industrial Informatics}, 2022.

\bibitem{wang2021transcd}
Z.~Wang, Y.~Zhang, L.~Luo, and N.~Wang.
\newblock Transcd: scene change detection via transformer-based architecture.
\newblock {\em Optics Express}, 29(25), 2021.

\bibitem{wang2022anodfdnet}
Z.~Wang, Y.~Zhang, L.~Luo, and N.~Wang.
\newblock Anodfdnet: A deep feature difference network for anomaly detection.
\newblock {\em arXiv preprint arXiv:2203.15195}, 2022.

\bibitem{wu2019detectron2}
Y.~Wu, A.~Kirillov, F.~Massa, W.-Y. Lo, and R.~Girshick.
\newblock Detectron2.
\newblock \url{https://github.com/facebookresearch/detectron2}, 2019.

\bibitem{xianjia2022analyzing}
Y.~Xianjia, S.~Salimpour, J.~{Pe\~na Queralta}, and T.~Westerlund.
\newblock Analyzing general-purpose deep-learning detection and segmentation
  models with images from a lidar as a camera sensor.
\newblock In {\em International Conference on Intelligent Systems Design and
  Engineering Applications (ISDEA), Lecture Notes in Electrical Engineering (to
  appear)}. Springer, 2022.

\bibitem{zavrtanik2021reconstruction}
V.~Zavrtanik, M.~Kristan, and D.~Sko{\v{c}}aj.
\newblock Reconstruction by inpainting for visual anomaly detection.
\newblock {\em Pattern Recognition}, 112, 2021.

\bibitem{zhang2022image}
W.~Zhang, X.~Tang, and J.~Zhang.
\newblock Image anomaly detection based on adaptive iteration and feature
  extraction in edge-cloud iot.
\newblock {\em Wireless Communications and Mobile Computing}, 2022, 2022.

\bibitem{zhang2016abnormal}
X.~Zhang, L.~Li, J.~Li, J.~Lyu, R.~Huang, and H.~Xing.
\newblock Abnormal appearance detection of substation based on image
  comparison.
\newblock In {\em MATEC Web of Conferences}, volume~59. EDP Sciences, 2016.

\bibitem{zhou2021novel}
Y.~Zhou, Y.~Song, S.~Cui, H.~Zhu, J.~Sun, and W.~Qin.
\newblock A novel change detection framework in urban area using multilevel
  matching feature and automatic sample extraction strategy.
\newblock {\em IEEE Journal of Selected Topics in Applied Earth Observations
  and Remote Sensing}, 14, 2021.

\end{thebibliography}

\end{document}